\pdfoutput=1
\PassOptionsToPackage{dvipsnames, svgnames, x11names}{xcolor} 

\documentclass[11pt]{article}

\usepackage[]{EACL2023}

\usepackage{times}
\usepackage{latexsym}

\usepackage[T1]{fontenc}

\usepackage[utf8]{inputenc}

\usepackage{microtype}

\usepackage{xspace}
\usepackage{inconsolata}
\usepackage{booktabs}
\usepackage{comment}
\usepackage{graphicx} 
\usepackage{tcolorbox} 
\usepackage{framed}
\usepackage{CJKutf8} 
\usepackage{multirow}

\def\figref#1{Figure~\ref{fig:#1}}
\def\figlabel#1{\label{fig:#1}\label{p:#1}}

\def\tabref#1{Table~\ref{tab:#1}}

\def\tablabel#1{\label{tab:#1}\label{p:#1}}

\def\secref#1{\S\ref{sec:#1}}
\def\seclabel#1{\label{sec:#1}}
\def\qref#1{Eq.~\ref{eqn:#1}}

\def\eqlabel#1{\label{eqn:#1}}

\def\method{\textsc{ToPro}\xspace}

\newcommand{\red}[1]{{\color{red} #1}}

\newcommand\blfootnote[1]{%
  \begingroup
  \renewcommand\thefootnote{}\footnote{#1}%
  \addtocounter{footnote}{-1}%
  \endgroup
}

\title{\includegraphics[scale=1]{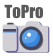} \method: Token-Level Prompt Decomposition for \\Cross-Lingual Sequence Labeling Tasks}

\vspace{3pt}
\author{
Bolei Ma$^\ast$$^\clubsuit$$^,$$^\heartsuit$,~
Ercong Nie$^\ast$$^\clubsuit$$^,$$^\heartsuit$,~
Shuzhou Yuan$^\spadesuit$,~
Helmut Schmid$^\clubsuit$,\vspace{3.5pt}\\
\textbf{
Michael Färber$^\spadesuit$,~
Frauke Kreuter$^\clubsuit$$^,$$^\heartsuit$$^,$$^\diamondsuit$
~and
Hinrich Schütze$^\clubsuit$$^,$$^\heartsuit$}\vspace{3.5pt}
\smallskip
\\
$^\clubsuit$LMU Munich, ~
$^\heartsuit$Munich Center for Machine Learning, \\
$^\spadesuit$Karlsruhe Institute of Technology,~
$^\diamondsuit$University of Maryland, College Park \smallskip\vspace{3.5pt}
\\
\smallskip
\texttt{bolei.ma@lmu.de,~nie@cis.lmu.de,~shuzhou.yuan@kit.edu}
}

\begin{document}
\begin{CJK}{UTF8}{gbsn}
\maketitle

\begin{abstract}
Prompt-based methods have been successfully applied to multilingual pretrained language models for zero-shot cross-lingual understanding.\blfootnote{$^\ast$ Equal contribution.}
However, most previous studies primarily focused on sentence-level classification tasks,
and only a few considered token-level labeling tasks such as Named Entity Recognition (NER) and Part-of-Speech (POS) tagging. 
In this paper, we propose \textbf{To}ken-Level \textbf{Pro}mpt Decomposition (\textbf{\method}), which facilitates the prompt-based method for token-level sequence labeling tasks. The \method method decomposes an input sentence into single tokens and applies one prompt template to each token. Our experiments on multilingual NER 
and POS tagging datasets demonstrate that \method-based fine-tuning outperforms Vanilla fine-tuning and Prompt-Tuning in zero-shot cross-lingual transfer, especially for languages that are typologically different from the source language English. Our method also attains state-of-the-art performance when employed with the mT5 model.
Besides, our exploratory study in multilingual large language models shows that \method performs much better than the current in-context learning method. 
Overall, the performance improvements show that \method could potentially serve as a novel and simple benchmarking method for sequence labeling tasks.\footnote{Our source code is available in this GitHub repository: \url{https://github.com/boleima/ToPro}.}

\end{abstract}

\section{Introduction}
\begin{figure}[ht]
    \centering
    \includegraphics[scale=0.925]{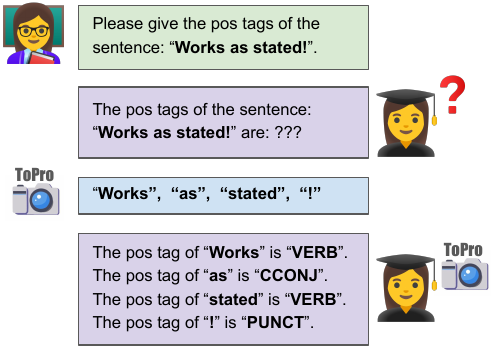}
    \caption{\method as a token-level prompting method for sequence labeling tasks. It decomposes the input sentence into single tokens and applies the prompt template to each token, inspired by human step-by-step logical thinking when solving this kind of task.}
    \figlabel{topro}
\end{figure}
As multilingual pretrained language models (MPLMs) continue to evolve~\citep{devlin-etal-2019-bert,conneau-etal-2020-unsupervised,liu-etal-2020-multilingual-denoising,xue-etal-2021-mt5,shliazhko2022mgpt}, zero-shot cross-lingual transfer methods are gaining increasing popularity within the multilingual NLP domain~\citep{lauscher-etal-2020-zero, nie-etal-2023-cross}. 
In light of the limited availability of 
training data in many low-resource languages, prior research~\citep{artetxe-etal-2020-cross, hu2020xtreme} employed zero-shot cross-lingual transfer learning by fine-tuning an MPLM on a high-resource language such as English, and then directly applying the fine-tuned system to low-resource languages.

Prompt-based learning~\citep{schick-schutze-2021-exploiting, schick-schutze-2021-shot, schick-schutze-2021-just} is steadily garnering traction in recent NLP research. 
Prompt-based methods reformulate downstream tasks as language modeling tasks by using prompts comprising a template and a set of label words. The prompt can be either discrete in a textual format or continuous, performing prompting directly in the embedding space of the model~\citep{liu2023pre}.
Much recent work highlights that applying prompt-based fine-tuning to MPLMs enables better zero-shot cross-lingual transfer performance~\citep{zhao-schutze-2021-discrete, huang-etal-2022-zero, nie-etal-2023-unleashing, zhou-etal-2023-enhancing}. 
However, they focus on sentence-level classification tasks such as sentiment analysis~\citep{keung-etal-2020-multilingual}, XNLI~\citep{conneau-etal-2018-xnli}, and paraphrase detection~\citep{zhang-etal-2019-paws}. 
Token-level sequence labeling tasks like Named Entity Recognition (NER) and Part-of-Speech (POS) Tagging rarely benefit from the advantages of prompt-based fine-tuning, primarily due to the intricate challenge of devising an appropriate prompt template.

To enhance the applicability of prompt-based learning to token-level sequence labeling tasks, we introduce the \textbf{To}ken-Level \textbf{Pro}mpt Decomposition (\textbf{\method}) method. \method splits the input sentence into tokens and creates a separate prompt for each token which asks for its label, following the human step-by-step logical thinking when solving these tasks, as shown in \figref{topro}. The evaluation on NER and POS tagging tasks shows that the \method-based fine-tuning achieves stronger zero-shot cross-lingual transfer performance than Vanilla fine-tuning and Prompt-Tuning, especially for languages that are typologically different from the source language (English).

Besides fine-tuning, in-context learning (ICL) is another common paradigm for applying large language models (LLMs) to downstream tasks. ICL prompts the LLM with few-shot demonstrations and/or natural language instructions to tackle a new task without updating any parameters~\citep{brown2020language, min-etal-2022-rethinking, wei2023larger}.
ICL using instructions without demonstrations has also been applied to zero-shot cross-lingual transfer with multilingual large language models (MLLMs) \citep{muennighoff-etal-2023-crosslingual, DBLP:conf/iclr/ShiSF0SVCTRZ0W23}. 
Nevertheless, akin to cross-lingual fine-tuning, existing methodologies encounter limitations when applied to sequence labeling tasks~\citep{ahuja2023mega, asai2023buffet}. 
To this end, we delve deeper into the integration of \method
with MLLMs and ascertain its effectiveness in an ICL paradigm.
Empirical findings substantiate the potential of \method as a benchmarking method for evaluating the performance of MLLMs in sequence labeling tasks.

To sum up, our contributions are as follows:
\\
(1) We propose a novel and simple method, called \method, which improves zero-shot cross-lingual transfer in token-level sequence labeling tasks by taking advantage of prompt-based learning. 
\\
(2) We substantiate the strength of \method in zero-shot cross-lingual fine-tuning through evaluations on NER and POS tagging tasks. Our method not only outperforms baselines for over 40 languages but also demonstrates efficacy in zero-shot English ICL, making it a promising benchmarking method for MLLMs in token-level tasks.
\\
(3) We conduct a thorough cross-lingual analysis, revealing that \method exhibits particularly strong performance for languages that are typologically different from the source language English.

\section{Related Work}

\paragraph{MPLMs and Cross-Lingual Transfer}
The progress in MPLMs has established them as the basis for cross-lingual transfer. MPLMs typically adopt the architecture of monolingual Transformer-based language models and
are pretrained on extensive unlabeled multilingual corpora.
Examples of MPLMs are mBERT~\citep{devlin-etal-2019-bert}, XLM-R~\citep{conneau-etal-2020-unsupervised}, mT5~\citep{xue-etal-2021-mt5}, Glot500~\citep{imanigooghari-etal-2023-glot500}, etc.
Empirical studies~\citep{karthikeyan2020cross, turc2021revisiting} have showcased the remarkable cross-lingual prowess of MPLMs which are fine-tuned on English training datasets,
and then used to predict on test datasets in other languages.
Several benchmark datasets such as XTREME~\citep{hu2020xtreme}, XTREME-R~\citep{ruder-etal-2021-xtreme}, and Taxi1500~\citep{ma2023taxi1500} have been created to assess the capabilities of multilingual models. 
The increasing popularity of prompt learning has drawn the attention of researchers
towards prompt-based methods for cross-lingual transfer~\citep{tu-etal-2022-prompt, ma2023promptbased}. 
Diverging from previous studies centered on sentence-level classification tasks, our work
applies prompt-based fine-tuning to token-level sequence labeling tasks.

\paragraph{MLLMs and In-Context Learning}
BLOOMZ and mT0~\citep{muennighoff-etal-2023-crosslingual} stand out as two representative multilingual models in the era of LLMs. 
Both are fine-tuned on the xP3 dataset which contains multi-lingual multi-task prompts.
BLOOMZ is built upon BLOOM~\citep{Scao2022BLOOMA1} while mT0 is built upon mT5~\citep{xue-etal-2021-mt5}.
\citet{brown2020language} demonstrated that LLMs like GPT-3 can acquire task-solving ICL abilities. 
The emergence of MLLMs opens up the possibility for conducting zero-shot cross-lingual ICL, as demonstrated by recent benchmarking efforts, for example MEGA~\citep{ahuja2023mega} and BUFFET~\citep{asai2023buffet}. 
However, current ICL methods using text-to-text prompting with a fixed output template for sequence labeling tasks ``\emph{consistently exhibit extremely poor performance}''~\citep{asai2023buffet} when applied to MLLMs, failing to
exploit their real cross-lingual transfer abilities. Contrary to that, our proposed \method method better reflects the potential of MLLMs on token-level tasks.

\paragraph{Prompt Methods for Sequence Labeling Tasks}
Although prompt-based methods proved useful in sentence-level classification tasks, they were seldom employed for token-level labeling tasks. 
\citet{cui-etal-2021-template} applied template-based prompting methods to the BART model~\citep{lewis-etal-2020-bart} for NER tasks. Their method is rank-based. They generate a sentence for each possible label and compute the probabilities of all generated sentences for the prediction, which can be expensive to decode.
\citet{ma-etal-2022-template} proposed a template-free prompting method for few-shot NER, called entity-oriented LM fine-tuning. However, they adopt the span-based task formulation of NER, resulting in more complexity, while our proposed method applies to NER tasks in the IOB (Inside-Outside-Beginning) tagging format. \citet{blevins-etal-2023-prompting} proposed a structural prompting method for sequence labeling tasks, which was adapted for multilingual benchmarking large language models in recent work~\citep{ahuja2023mega}.

\section{\method for Fine-Tuning}

\paragraph{Problem Formulation}
In prompt-based learning, there is a pattern-verbalizer pair (PVP) \citep{schick-schutze-2021-exploiting} consisting of (i) a \emph{prompt pattern} which converts the input text into a cloze-style question with a mask token, and (ii) a \emph{verbalizer} which maps the labels
onto representative words from the LM's vocabulary. This aligns well with the nature of text classification tasks where one label is predicted based on the input text. As \figref{prompt_cla} shows, the input text $X$ of a sentiment analysis task can be reformulated with a prompt pattern $P(\cdot)$ into a prompted input representation $P(X) = $ ``\colorbox{LightSteelBlue1}{Works as stated!} \colorbox{Moccasin}{In summary, the product was [MASK].}'' 
\begin{figure}[ht]
    \centering
    \includegraphics[scale=0.7]{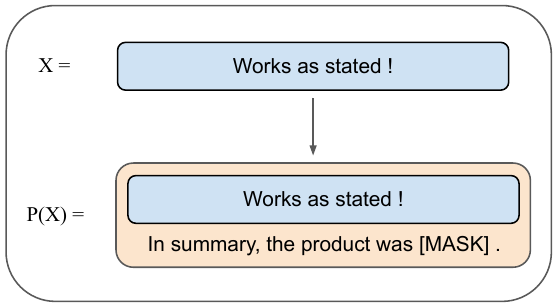}
    \caption{A prompt example for text classification.}
    \figlabel{prompt_cla}
\end{figure}
The prompt $P(X)$ is processed by the LM to determine the most likely
verbalizer word in the masked position. The label corresponding to this
verbalizer is the prediction which is evaluated against the gold
standard.

However, in sequence labeling tasks, each token of the input should receive a label. Thus, it is not possible to apply this type of prompt pattern with one mask token directly for token classification.

\paragraph{Token-Level Prompt Decomposition (\method)}
When given such a token-level sequence labeling task, 
a human usually solves the task token by token. 
Inspired by this human process as well as the prompt design for sentence classification tasks,
we propose a new prompting method \method for token classification which decomposes an input sentence into tokens and generates a series of prompts -- one prompt for each token. 
Let $X = x_1, x_2, ... , x_m$ denote an input sentence consisting
of $m$ tokens.  Our prompt generator function $P(T,X)$
generates $m$ prompts by filling the template
$T(\cdot,\cdot)$ with the sentence $X$ and each of the
tokens $x_1, x_2,...,x_m$, respectively.
\begin{equation}
  \eqlabel{topro}
  P(T,X) = \{T(X,x_1), ..., T(X,x_m)\}
\end{equation}

\figref{prompt_label} shows the prompts generated by $P(T,X)$ for the input $X=$
``Works as stated !'' and the template $T(X,x_i)=$
``\colorbox{LightSteelBlue1}{$X$} \colorbox{Moccasin}{The POS
  tag of} \colorbox{Lavender}{$x_i$} \colorbox{Moccasin}{is a kind of
  [MASK]}.''

\begin{figure}[ht]
    \centering
    \includegraphics[scale=0.57]{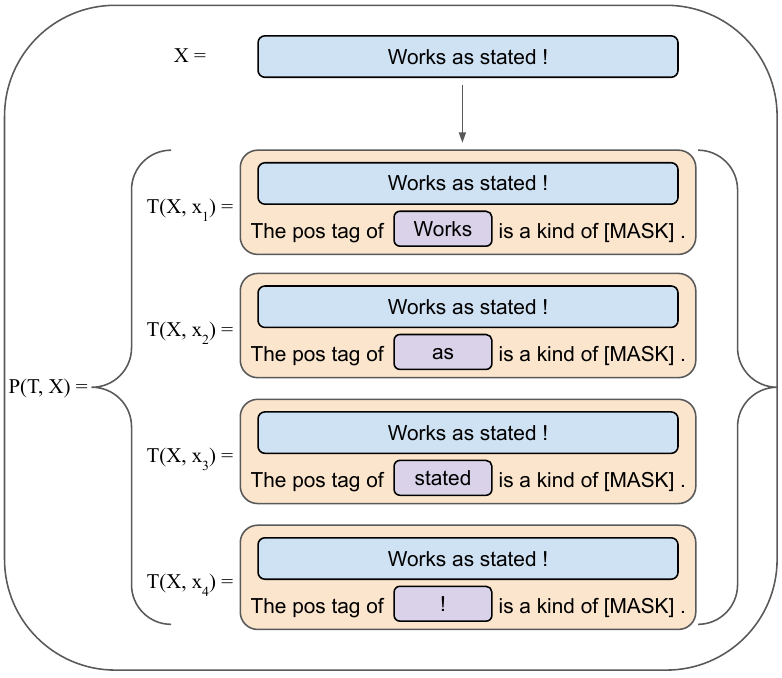}
    \caption{An example of \method framework for sequence labeling.}
    \figlabel{prompt_label}
\end{figure}

\paragraph{Prompt-Based Fine-Tuning and Cross-Lingual Transfer} Following \citet{ma2023promptbased}, we conduct prompt-based fine-tuning to evaluate our \method approach in a zero-shot cross-lingual context.
Let $D = \{(X_1, Y_1), ... , (X_n, Y_n)\}$ denote the set of training examples in the source language, where $X_1,...,X_n$ are token sequences and $Y_1,...,Y_n$ are tag sequences. 
Given $(X, Y) \in D$, the \method function $P(T,X)$ reformulates the input sentence $X$ into a set of cloze-style questions $\{T(X,x_1),...,T(X,x_m)\}$ with masked tokens. The pretrained language model $M$ with trainable parameters $\theta$ performs masked token prediction and returns the probabilities $p(\cdot) = M(T(X,x_i),\theta)$ of all candidate words for the masked token in the prompt $T(X,x_i)$. The verbalizer function $V(\cdot)$ is a bijective mapping from the set of class labels $L$ to a set of verbalizers from the source language vocabulary. For each token, we predict the tag $\hat{y}$ whose verbalizer $V(\hat{y})$ receives the highest probability from model $M$:
\begin{equation}
\eqlabel{argmax}
\hat{y} = \arg\max_{y \in L} p(V(y))
\end{equation} 
We fine-tune the parameters $\theta$ of model $M$ by minimizing the cross-entropy loss function $\ell(D,\theta)$: 
\begin{equation}
\small
  \ell(D,\theta) = -\sum_{(X,Y) \in D} \sum_{i=1}^{|Y|} \log M(T(X,x_i), \theta)(V(y_i))
\end{equation} 
The fine-tuned model is used to predict the labels of the target language examples
$\{X_1^{\prime}, ... , X_n^{\prime}\}$ 
using the same prompt pattern $T(\cdot,\cdot)$ and verbalizer $V(\cdot)$ as during fine-tuning . The best tags $Y_j'$ for each example $X_j'$ are predicted according to \qref{argmax}.


\section{Experimental Setups}
\subsection{Datasets and Prompt Designs}
We choose the following two representative datasets for sequence labeling tasks:

\paragraph{PAN-X}
\citep{pan-etal-2017-cross}, also called WikiANN, is a multilingual NER dataset based on Wikipedia articles including 282 languages. 
In our work, we use the subset of 48 languages which is part of the XTREME benchmark~\citep{hu2020xtreme} to facilitate comparisons with related work.

For each token $x_i$ of an input sequence $X$, we use the following prompt template $T(X,x_i)$
:

\begin{tcolorbox}
    [colback=gray!20, colframe=gray!100, sharp corners, leftrule={3pt}, rightrule={0pt}, toprule={0pt}, bottomrule={0pt}, left={2pt}, right={2pt}, top={3pt}, bottom={3pt}]
{\small
$T(X,x_i) = X \ \circ$ `` The named entity of '' $\circ \ x_i \ \circ$ `` is a kind of: [MASK].''}
\end{tcolorbox}

The PAN-X dataset is annotated with location (LOC), person (PER), and organization (ORG) in IOB2 format. These labels are difficult to understand for the language model. Therefore, we replace them with real words and train the model to predict those instead. However, the model can only predict single words from its vocabulary as fillers for the [MASK] position. So, we choose the replacement words from the model's vocabulary.

As IOB2 annotates the beginning of a name and its remaining tokens with different tags, we use a word and its hyponym to represent the beginning of a name and its remaining tokens, respectively.
For instance, we use the hypernym “location” for the beginning of the LOC and the hyponym “place” for the other words which should be semantically inside of the term “location”. The verbalizer function $V(\cdot)$ for tag set $Y$ is defined as follows:

\begin{tcolorbox}
  [colback=gray!20, colframe=gray!100, sharp corners, leftrule={3pt}, rightrule={0pt}, toprule={0pt}, bottomrule={0pt}, left={2pt}, right={2pt}, top={3pt}, bottom={3pt}]
  {\small
    \begin{tabbing}
      V(B-ORG) = organization ~ \= V(I-ORG) = body\kill
      V(B-LOC) = location \> V(I-LOC) = place\\
      V(B-ORG) = organization \> V(I-ORG) = body\\
      V(B-PER) = person \> V(I-PER) = name\\
      V(O) = other
    \end{tabbing}
  }
\end{tcolorbox}

\paragraph{UDPOS} This dataset was extracted from the Universal Dependency treebanks \citep{ud}. It contains 38 languages and is part of the XTREME benchmark~\citep{hu2020xtreme}.

Similarly as for the PAN-X dataset, we use the following prompt template $T(X,x_i)$ for token $x_i$ of an input sequence $X$ by paraphrasing the tags with semantically related words:

\begin{tcolorbox}
    [colback=gray!20, colframe=gray!100, sharp corners, leftrule={3pt}, rightrule={0pt}, toprule={0pt}, bottomrule={0pt}, left={2pt}, right={2pt}, top={3pt}, bottom={3pt}]
{\small 
$T(X,x_i) = X \ \circ$ `` The pos tag of '' $\circ \ x_i \ \circ$ `` is a kind of: [MASK].''}
\end{tcolorbox}

The dataset has 14 labels.  A detailed description of the tags can be
found in Table \ref{tab:postag} of the Appendix.

We define the verbalizer $V(\cdot)$ for the 14 tags as follows: 

\begin{tcolorbox}
    [colback=gray!20, colframe=gray!100, sharp corners, leftrule={3pt}, rightrule={0pt}, toprule={0pt}, bottomrule={0pt}, left={2pt}, right={2pt}, top={3pt}, bottom={3pt}]
    {\small
      \begin{tabbing}
        V(SCONJ) = condition \= V(SCONJ) = condition \kill
        V(ADJ) = modification \> V(ADP) = position \\
        V(ADV) = verbal \> V(AUX) = auxiliar \\
        V(CCONJ) = link \> V(DET) = determine \\
        V(INTJ) = mode \> V(NOUN) = thing \\
        V(NUM) = number \> V(PART) = functional \\
        V(PRON) = reference \> V(PROPN) = name \\
        V(PUNCT) = punct \> V(SCONJ) = condition \\
        V(SYM) = symbol \> V(VERB) = verb \\
        V(X) = other
      \end{tabbing}
}
\end{tcolorbox}

We cannot select words like “adjective” and “adverb” which would better
  represent the meanings of the tags because the verbalizers have to
  come from the vocabulary of the PLM so that the masked language model
  is able to predict them as a single unit. Instead, we use semantically
  related words from the vocabulary as verbalizers.

\subsection{Baselines}
We compare our approach with the following baselines: 

\paragraph{Vanilla Fine-Tuning (Vanilla)} The vanilla fine-tuning method predicts the token labels through the hidden embeddings of each token in the output layer without using a prompt pattern. We use the cross-entropy loss as the objective function for fine-tuning and \texttt{AdamW} for optimization with a learning rate of 1e-5. The fine-tuned models are used to predict the test data.

\paragraph{Prompt-Tuning (PT)} Prompt-Tuning
only trains a small number of parameters, e.g., a continuous prompt or a task classifier \cite{lester-etal-2021-power, liu-etal-2022-p}. We implement the prompt-tuning method of \citet{tu-etal-2022-prompt} for zero-shot cross-lingual transfer by tuning the prefix prompts and layer prompts for the two sequence labeling tasks.

\subsection{Multilingual Models}
The following MPLMs from the HuggingFace Transformers library \citep{wolf-etal-2020-transformers} are applied in our main experiments: 

\begin{itemize}
\item \textbf{Encoder-Only Models:} including the multilingual BERT model \citep{devlin-etal-2019-bert} \texttt{bert-base-multilingual-cased} (B) and the XLM-R model \citep{conneau-etal-2020-unsupervised} \texttt{xlm-roberta-base} (X), and

\item \textbf{Encoder-Decoder Model:} the multilingual T5 model \citep{xue-etal-2021-mt5} \texttt{mt5-base} (T). We also include the encoder-decoder model mT5 in our experiments, as we wish to explore the potential of \method with different types of models. 
Details on how we trained mT5 for sequence labeling tasks
are shown in \secref{mt5_training}. 
\end{itemize}


\section{Results and Analysis}
\subsection{Main Results}

\tabref{overview_sequence} gives an overview of the average results\footnote{Since we are interested in the zero-shot cross-lingual transfer performance, we do not include the English results in the average performance. Our evaluation metric is the weighted average F1-score.} 
on PAN-X and UDPOS. We find that \method Fine-Tuning outperforms Vanilla Fine-Tuning and Prompt-Tuning obviously on both tasks in mBERT and XLM-R settings: On PAN-X, the performance is improved by \textbf{19.18\%} and \textbf{25.16\%} compared to Vanilla and Prompt-Tuning respectively when trained with mBERT, and by \textbf{18.73\%} and \textbf{26.98\%} with XLM-R. On PAN-X, the performance is improved by \textbf{5.27\%} and \textbf{6.24\%} compared to Vanilla and Prompt-Tuning respectively when trained with mBERT, and by \textbf{3.74\%} and \textbf{4.3\%} with XLM-R. 

In the mT5 setting, the \method Fine-Tuning outperforms Vanilla Fine-Tuning on both tasks as well, namely by \textbf{28.63\%} on PAN-X, and by \textbf{14.72\%} on UDPOS. We notice that the mT5 model performs even better than the two encoder-only models and achieves SOTA performance\footnote{Based on the evaluation results available at \url{https://sites.research.google/xtreme/dataset}, as of Jan. 23, 2024, the SOTA performance in structured prediction, calculated as the mean value of PANX-X and UDPOS, is 84.6. Our mT5 model, when used with \method, achieves an impressive score of 89.47.}, showing the potential of \method with different model types. We find that Prompt-Tuning does not work well with mT5, as it requires more training epochs for the model to achieve subtle performance improvements, necessitating even longer training time compared to the Vanilla baselines. One possible reason for this could be the limited number of trainable parameters in mT5 with Prompt-Tuning, as only 0.002\% of the parameters are updated with our current prompt settings (see \secref{mt5_training}). We exclude the results of Prompt-Tuning for mT5 because the increased training resources do not align with the efficiency-focused goals of Prompt-Tuning as a training methodology.

When comparing performances on the two tasks generally, we notice that the performance shows greater improvement on PAN-X with all three models, indicating that the PAN-X for NER task has a greater improvement potential. 

\begin{table}[ht]
\renewcommand\arraystretch{1.2}
\setlength\tabcolsep{6pt}
\centering
\footnotesize
\begin{tabular}{llcc}
\toprule
 Model & Method & PAN-X & UDPOS  \\
 \midrule
 \multirow{3}*{mBERT} &Vanilla Fine-Tuning & 62.73& 70.89 \\
 &Prompt-Tuning& 56.76& 69.91\\
 &\method Fine-Tuning & \textbf{81.91} & \textbf{76.16}\\
 \midrule
 \multirow{3}*{XLM-R} &Vanilla Fine-Tuning & 61.30 &72.42\\
   &Prompt-Tuning& 53.05& 71.86\\
 &\method Fine-Tuning & \textbf{80.03} &\textbf{76.16} \\ 
  \midrule
 \multirow{3}*{mT5} &Vanilla Fine-Tuning & 64.19 &71.38\\
   &Prompt-Tuning& -* & -* \\
 &\method Fine-Tuning & \textbf{92.82} & \textbf{86.11}\\ 

\bottomrule
\end{tabular}
\caption{Overview of average results on PAN-X and UDPOS. $*$: The results of PT with mT5 are excluded from the comparison as the F1 scores are 0 for
the current parameter settings.
}
\tablabel{overview_sequence}
\end{table}

\begin{table}[ht]
    \centering
    \includegraphics[scale=0.557]{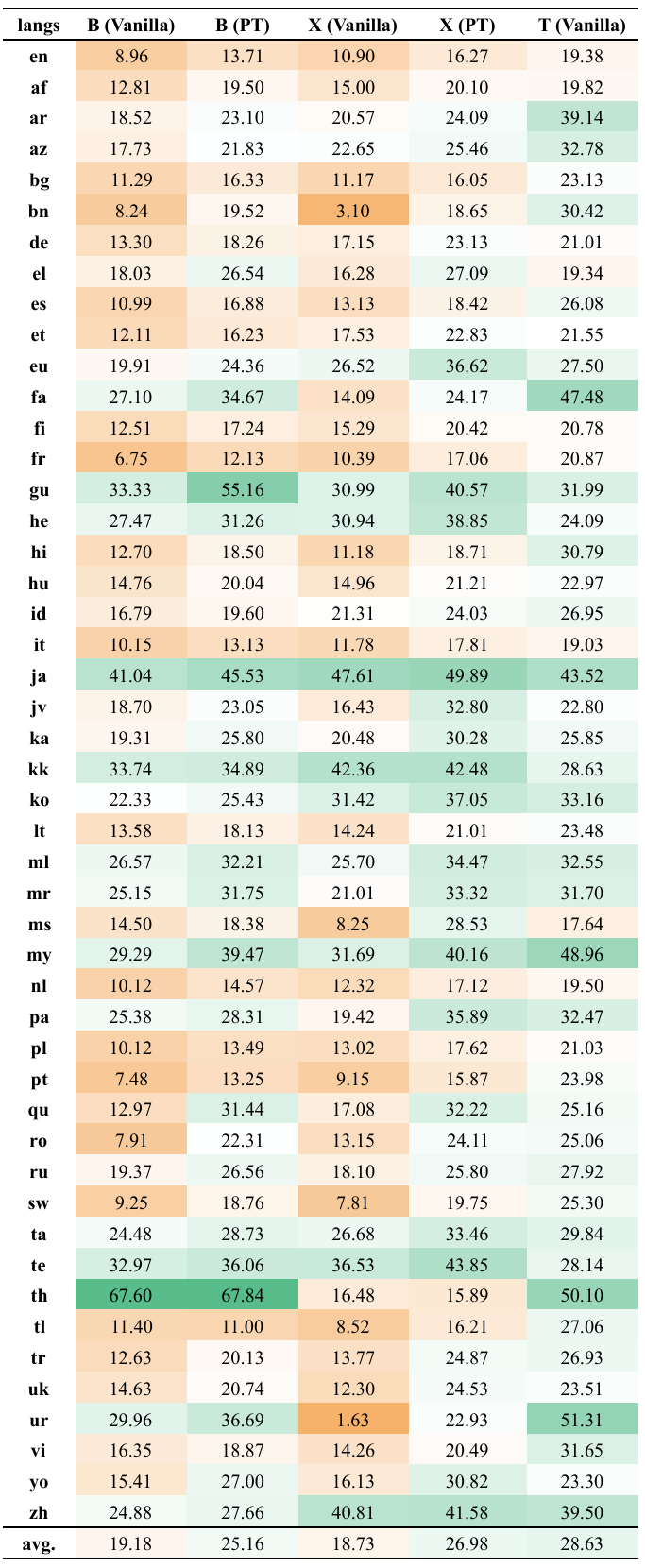}
    \caption{Performance difference ($-$) of \method to Vanilla or Prompt Tuning (PT) with mBERT (B), XLM-R (X) and mT5 (T) on PAN-X.}
    \tablabel{panx_dif}
\end{table}

\subsection{Cross-Lingual Transfer Analysis}
\seclabel{cross-lingual}

The detailed results of the cross-lingual transfer performance of \method compared to the baselines for each target language are documented in \tabref{panx_full} and \tabref{udpos_full} in \secref{detailed_results} of the appendix.
\tabref{panx_dif} and \tabref{udpos_dif} above show the performance
improvements of \method compared to the baselines for each language.
Overall, \method-based Fine-Tuning outperforms Vanilla Fine-Tuning and Prompt-Tuning on average. However, we can notice individual performance differences between the languages.

\begin{table}[ht]
    \centering
    \includegraphics[scale=0.62]{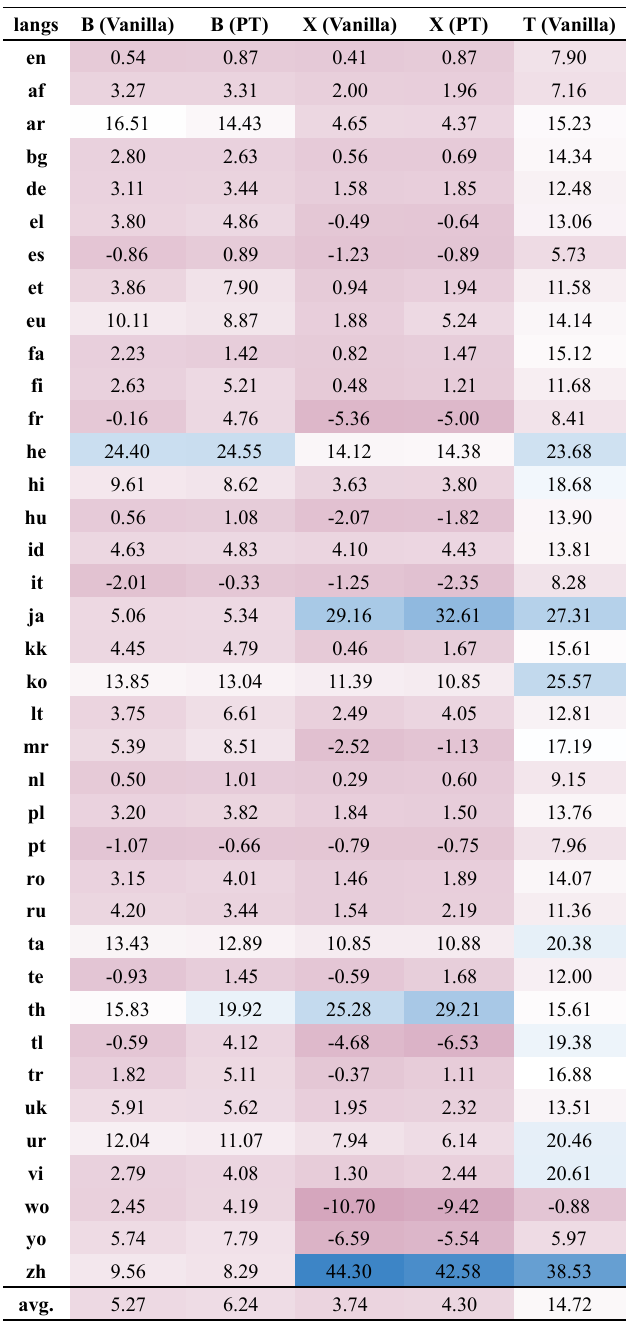}
    \caption{Performance difference ($-$) of \method to Vanilla or PT with mBERT (B), XLM-R (X) and mT5 (T) on UDPOS.}
    \tablabel{udpos_dif}
\end{table}

On both tasks, we find that the performance gain of \method for English (en) is among the lowest across all languages. 
Since English is the language on which the models have been fine-tuned, we conclude that \method is particularly effective in cross-lingual zero-shot scenarios.
The reason could be that the models are only fine-tuned on the English dataset. Therefore, 
 {\method}'s potential performance improvement is smaller for English than for other languages.


\begin{table*}[ht]
\renewcommand\arraystretch{1.35}
\setlength\tabcolsep{3pt}
\centering
\scriptsize
\begin{tabular}{l}
\toprule
\textbf{Input sequence \& Its Gloss \& Tags (True, Vanilla, \method) \& Translation}\\
\midrule
\textbf{Case 1 (zh)} \\
\textbf{Input:} \begin{CJK}{UTF8}{bsmi} { 溫暖 海岸 \red{是 指} 西班牙 穆爾西亞 自治 區 \red{長 約} 250 公里 的 地中 海 \red{海岸 線} 。}\end{CJK} \\
\textbf{Gloss:}  Warm coast \red{be refer} Spain Murcia autonomous region \red{long approximately} 250 km 's Mediterranean sea coast line .\\
\textbf{True:}  propn noun  \red{verb verb} propn propn verb part \red{adj adv} num noun part propn part \red{noun part} punct \\
\textbf{Vanilla:}  propn punct  \red{punct punct} punct propn punct punct \red{punct punct} num num punct noun noun \red{punct punct} punct (0.19 F1)\\
\textbf{\method:}  propn propn  \red{aux verb} propn propn propn propn \red{adj adv} num noun adp noun noun \red{propn noun} punct (0.47 F1)\\
\textbf{Translation:}  The Costa Cálida is the 250-kilometer-long Mediterranean coastline of the Autonomous Region of Murcia, Spain.\\

\midrule
\textbf{Case 2 (ja)} \\
\textbf{Input:} \begin{CJK}{UTF8}{min}\red{政府} が \red{もっと} 宗教法人 の 定義 を 厳しく \red{し} , こういう 団体 は \red{排除 す} べき 。 \end{CJK}\\
\textbf{Gloss:} \begin{CJK}{UTF8}{min}\red{government} subject \red{more} religious organisation of definition object strictly \red{that} , such association topic \red{exclude do} must . \end{CJK}\\
\textbf{True:} \red{noun} adp \red{adv} noun adp noun adp adj \red{aux} punct adj noun adp \red{verb aux} aux punct  \\
\textbf{Vanilla:} \red{det} punct \red{punct} punct punct noun punct punct \red{punct} punct punct punct punct verb \red{punct punct} punct \ (0.29 F1)\\
\textbf{\method}: \red{noun} part \red{adv} noun part noun pron adj \red{verb} punct adj noun pron \red{verb verb} adj punct \ (0.64 F1)\\
\textbf{Translation:} ``The government should tighten the definition of religious corporations and eliminate such organizations.''\\

\midrule
\textbf{Case 3 (de)} \\
\textbf{Input:} ,,  Mich interessiert etwas , wenn es mich \red{zu} Teilnahme \red{zu} erregen weiß ``, sagte er in einem deutschen Fernsehen . \\
\textbf{Gloss:} `` me interest something , if it me \red{to} attendance \red{to} irritate knows '' , said he in a German television .\\
\textbf{True:} punct pron verb pron punct sconj pron pron \red{adp} noun \red{part} verb verb punct punct verb pron adp det adj noun punct  \\
\textbf{Vanilla:} punct pron verb pron punct sconj pron pron \red{adp} noun \red{part} verb verb punct punct verb pron adp det adj noun punct  \ (1.00 F1)\\
\textbf{\method:} punct pron verb pron punct sconj pron pron \red{part} noun \red{part} verb verb punct punct verb pron adp det adj noun punct  \ (0.95 F1)\\
\textbf{Translation:} ``I am interested in something if it knows how to excite me to participate'', he said on German television.''\\

\midrule
\textbf{Case 4 (nl)} \\
\textbf{Input:} ,, We hebben een concept \red{nodig} voor verandering ``, zei Djindjic in een interview met de Duitse televisie . \\
\textbf{Gloss:} `` we have a concept \red{necessary} for change '' , said Djindjic in a interview with the German television .\\
\textbf{True:} punct pron verb det noun \red{adj} adp noun punct punct verb propn adp det noun adp det adj noun punct  \\
\textbf{Vanilla:} punct pron verb det noun \red{verb} adp noun punct punct verb propn adp det noun adp det adj noun punct \ (0.95 F1)\\
\textbf{\method:} punct pron verb det noun \red{verb} adp noun punct punct verb propn adp det noun adp det adj noun punct  \ (0.95 F1)\\
\textbf{Translation:} ``We need a concept for change'', Djindjic said in an interview with German television.\\

\bottomrule
\end{tabular}
\caption{
  Comparison of the output of \method and Vanilla for selected UDPOS examples with XLM-R. The interesting tokens and their tags are marked \red{red}. The sentences were translated into English using \url{www.deepl.com}.}
\tablabel{error}
\end{table*}

On \colorbox{DarkSeaGreen2}{PAN-X}, \method outperforms Vanilla and Prompt-Tuning across all target languages, with some language-independent variations. The improvements in languages such as Persian (fa), Gujarati (gu), Hebrew (he), Japanese (ja), Kazakh (kk), Burmese (my), Telugu (te), Thai (th), Urdu (ur), and Chinese (zh) are above the average. 
All these languages are from different language groups to English and have different writing systems. We can conclude that the performance improvement of \method is particularly high for languages that differ a lot from English, further indicating the cross-lingual ability of our prompt-based method.

On \colorbox{LightSteelBlue1}{UDPOS}, \method outperforms Vanilla and PT in most of the languages, although there are some languages for which \method performs slightly worse and the overall performance gain is not as high as on PAN-X. Typically, the improvements for languages such as Arabic (ar), Basque (eu), Hebrew (he), Korean (ko), Tamil (ta), Thai (th), Urdu (ur), and Chinese (zh) are above average. The improvements over Vanilla in Chinese reach 44.3\% and 38.53\% for XLM-R and mT5, respectively, and the improvement over PT in Chinese is 42.58\%.

Overall, the results show that \method outperforms Vanilla and PT on both sequence labeling tasks, indicating that the \method method has a better ability to transfer knowledge cross-lingually.
And the NER performance is even better than the performance for POS tagging. 
When analyzing the performances for individual languages, we find that \method has a
strong performance for zero-shot cross-lingual transfer, particularly in languages with low similarity to English and different writing systems. The prompt-based approach seems to mitigate the language barriers and facilitate cross-lingual transfer. Additionally, the results vary across target languages, highlighting the importance of language typology and writing systems in determining the effectiveness of \method.

\subsection{Error Analysis} 
\seclabel{error_analysis}
In this section, we analyze selected instances from the UDPOS task with
typical annotation errors by the models in \tabref{error}.

The first example is a sentence in Chinese (zh), which is typologically and orthographically quite different from the training language English. The first two tokens marked red in this example \red{是} (``be'') \red{指} (``refer to'') are a pair of verbs, a so-called double-verb structure. They are both predicted by Vanilla as \texttt{PUNCT} (punctuation), but by \method as \texttt{AUX} (auxiliary) and \texttt{VERB} which are quite close to the corrected tags, as the auxiliary itself is a special kind of verb. The tokens \begin{CJK}{UTF8}{bsmi}\red{長} (``long'') \red{約} (``around'')\end{CJK} are predicted by Vanilla again as \texttt{PUNCT}, but correctly by \method as \texttt{ADJ} and \texttt{ADV}. Moreover, the tokens \begin{CJK}{UTF8}{bsmi}\red{海岸} (``coast'') \red{線} (``line'')\end{CJK} are predicted by Vanilla still as \texttt{PUNCT}, and by \method as \texttt{PROPN} (proper noun) and \texttt{NOUN}, which, though not the same as the original tags \texttt{NOUN} and \texttt{PART} (particle), are already close to the original tags as they are all a kind of noun. In this case, we notice that the Vanilla model tends to predict \texttt{PUNCT} for the majority of the tokens, whereas the \method method often predicts the correct tags or at least semantically related tags. 

The second example is in Japanese (ja), which is also typologically and orthographically quite different from English. The first token \begin{CJK}{UTF8}{min}\red{政府}\end{CJK} (``government'') is predicted by Vanilla as \texttt{DET} (determiner) which is somehow close to its original tag, and correctly by \method as \texttt{NOUN}. The token \begin{CJK}{UTF8}{min}\red{もっと}\end{CJK} (``more'') is predicted by Vanilla as \texttt{PUNCT}, but correctly by \method as \texttt{ADV}. The token \begin{CJK}{UTF8}{min}\red{し}\end{CJK} (``that'') is originally \texttt{AUX}, and it is predicted by Vanilla again as \texttt{PUNCT}, but by \method as \texttt{VERB}, which is already close to the meaning of \texttt{AUX}. And the token pair \begin{CJK}{UTF8}{min}\red{排除} (``exclude'') \red{す} (``do'')\end{CJK} has original labels \texttt{VERB AUX}, and is predicted by \method as \texttt{VERB VERB}, which are still very close to their original labels. However, Vanilla predicts them again as \texttt{PUNCT PUNCT}. Similar to the Chinese example, the Vanilla method tends to predict \texttt{PUNCT} for unfamiliar tokens, whereas \method generates the correct tags or at least tags close to the correct tags.

The third example is an input sentence in German, which is very close to the English language. We reach therefore very high F1 scores both with Vanilla and \method approaches. Noticeably, in this example, there is one token \red{zu} (``to'') with two different kinds of POS: \texttt{ADP} and \texttt{PART}. The Vanilla correctly detects the difference, while \method classifies both tokens as \texttt{PART}.
This is a shortcoming of TOPRO's token-wise prompting strategy which
generates identical prompts for both occurrences of ``zu''.

The fourth example is a Dutch (nl) input sentence. Dutch is also closely related to English. We reach a high F1 score of 0.95 with both Vanilla and \method approaches. The two approaches make the same error by predicting the token \red{nodig} (``necessary'') as \texttt{VERB}, which should be \texttt{ADJ}.

In conclusion, the first two examples show that \method works much better than Vanilla for languages that are typologically different from the source language of training. And even when predicting false POS tags, \method tends to predict tags semantically close to the correct tags. The last two examples show the slightly worse performance of \method for languages that are close to the source language of training. These findings support our claim in \secref{cross-lingual}.

\subsection{Exploratory Study in MLLMs} 

Previous benchmarking work on MLLMs has encountered difficulties in the sequence labeling tasks.  The BUFFET work by \citet{asai2023buffet} evaluates the multilingual capability of large generative language models across a wide variety of tasks. The benchmark contains NER tasks using the PAN-X test dataset. They employ a text-to-text prompting method to define the prompts for sequence labeling tasks. 
In this method, a model extracts all named
entities with named entity tags from the input sentences.
The target output string has to follow a fixed template. If the actual output produced by the system during evaluation does not exactly follow this format, it cannot be evaluated against the gold standard.

As \tabref{llm} shows, the text-to-text prompting approach works badly for MLLMs. The BUFFET work evaluated with the MLLMs BLOOMZ and mT0, but achieved very poor performance. The performance of the mT0 model was reported as $0.0$, suggesting that this prompting approach fails to work for mT0. Therefore, such text-to-text prompting as a benchmarking method cannot properly reflect the cross-lingual capability of currently existing MLLMs for sequence labeling tasks. 

\begin{table}[ht]
\renewcommand\arraystretch{1.45}
\setlength\tabcolsep{6pt}
\centering
\small
\begin{tabular}{l c c}
\toprule
~ & BUFFET & \method (Ours) \\
\midrule
\texttt{bloomz-7b1} & 7.6* & 13.98 \\
\texttt{mt0-xxl} & 0.0* & 18.09 \\
\bottomrule
\end{tabular}
\caption{Results of \texttt{bloomz-7b1} and \texttt{mt0-xxl} models on the PAN-X dataset. $*$: Values were directly extracted from the original paper.}
\tablabel{llm}
\end{table}

In contrast, 
our proposed \method prompting method performs much better as Table 5 shows.
We apply the \method prompt to MLLMs in a zero-shot ICL paradigm without parameter updating. 
The results of our exploratory study in MLLMs indicate that, compared to the current commonly used prompting method in MLLM benchmarking work, \method is a more promising prompting one for MLLMs in sequence labeling tasks.
Detailed information on the experimental settings and results regarding \method applied to MLLMs is provided in \secref{llm_details}.

\section{Conclusion}
In our work, we introduce \method for token-level sequence labeling tasks, a novel and simple method that adopts the basic framework of prompting from sentence classification tasks and applies the prompt template to each token in a sentence.
We evaluate the \method-based fine-tuning for zero-shot cross-lingual transfer
and compare it to Vanilla fine-tuning and Prompt-Tuning baselines.
We apply \method with three MPLMs on two representative sequence labeling tasks: NER and POS tagging. Our experiments show that \method outperforms the baselines with the MPLMs and achieves SOTA performance with mT5. We further discovered that the performance improvement of \method is generally more obvious in the cross-lingual context, especially for languages that are linguistically very different from the source language English, highlighting its cross-lingual ability. Additionally, we applied the \method method
to MLLMs and noticed better performances of \method as well, compared to existing benchmarking work. 
Overall, \method shows a noticeable performance improvement and could serve as a potential benchmark for sequence labeling tasks for future studies in prompt-based learning.

\section*{Limitations}
Our method has a shortcoming which we observed in the third error
example in \secref{error_analysis}: \method generates identical
prompts when a token occurs multiple times in a sentence. This results
in errors unless all occurrences have the same goldstandard label.  We
plan to explore in future work whether marking the currently annotated
token in the input sentence solves this problem.

While 
\method outperforms the baseline methods, its training takes much longer because the model is prompted with each token. For example, the Vanilla fine-tuning on PANX with mBERT takes around 30 minutes for 5 epochs, while the \method fine-tuning takes around 125 minutes, i.e. training is 4 times slower. Future work could pay more attention to improving the efficiency of the method.

Furthermore, the current study only considers prompt patterns and verbalizers that have been manually designed by the authors. Future work could focus on methods that automatically find a suitable prompt. Also, dynamic prompt applications could be taken into account, to look for a best-performing prompt.

\section*{Ethics Statement}
This research was conducted in accordance with the ACM Code of Ethics. 
Both datasets \cite{pan-etal-2017-cross, ud} that we use are publicly available. 
We report only aggregated results in the main paper.
We have not intended or do not intend to share any Personally Identifiable Data with this paper.

\section*{Acknowledgements}
We want to thank the anonymous reviewers for their invaluable contributions and constructive feedback. This work was partially supported by funding from BERD@NFDI, Munich Center for Machine Learning (MCML), and China Scholarship Council (CSC). 

Besides, we express our heartfelt gratitude to Mengmeng Xu for her efforts in designing the \method logo:  \includegraphics[scale=1]{fig/topro_symb.pdf}.

\bibliography{custom, anthology}
\bibliographystyle{acl_natbib}

\vspace{5pt}
\appendix
\section{Appendix}
\seclabel{appendix}

\subsection{Hyperparameter Settings}
\seclabel{hyperparameter_settings}

To avoid random effects on training, we trained each experiment with 5 different random seeds $\{10, 42, 421, 510, 1218\}$ and took the average results. The applied hyperparameters for the three models are documented in \tabref{hyperparams}. As Prompt-Tuning only tunes a few parameters, we increase the training epochs by 10 times more than Vanilla and \method Fine-Tuning.

\begin{table}[ht]
  \centering
  \small
  \renewcommand\arraystretch{1.45}
    \setlength\tabcolsep{7pt}
  \begin{tabular}{lcc}
    \toprule
    Hyperparameter & B \& X& T \\
    \midrule
    \texttt{epochs} &    5 (PT: 50)& 10 (PT: 100)\\
    \texttt{learning\_rate}   & 1e-5& 3e-5\\
    \texttt{batch\_size} &8& 24\\
    \texttt{grad\_acc\_steps} &4&4\\
    \texttt{max\_seq\_length}  &128&128\\
    \texttt{max\_target\_length}  &-&150\\
    \texttt{num\_beam\_search}  &-&3\\
    \bottomrule
  \end{tabular}
\caption{Hyperparameters}
\tablabel{hyperparams}
\end{table}

\subsection{mT5 Training} 
\seclabel{mt5_training}
In order to align with the text-to-text transformer format, we have redefined the output structure for both NER and POS tasks, drawing inspiration from the prior work of mT5 \cite{xue-etal-2021-mt5}. For the input text, we introduce task descriptions as prompts, specifically ``NER tagging:'' for the PAN-X dataset and ``POS tagging:'' for the UDPOS dataset. Regarding the target text, we append tags to each token and insert delimiters between tokens to create a coherent sequence of text. The following example illustrates our preprocessing procedure using a sample from the UDPOS dataset for Vanilla fine-tuning:

\begin{tcolorbox}
    [colback=gray!20, colframe=gray!100, sharp corners, leftrule={3pt}, rightrule={0pt}, toprule={0pt}, bottomrule={0pt}, left={2pt}, right={2pt}, top={3pt}, bottom={3pt}]
{\small
\begin{itemize}
\item \textbf{Input text:} POS tagging: On the other hand, it looks pretty cool .
\item \textbf{Target text:} ADP: On \$\$ DET: the \$\$ ADJ: other \$\$ NOUN: hand \$\$ PUNT: , \$\$ PRON: it \$\$ VERB: looks \$\$ ADV: pretty \$\$ ADJ: cool \$\$ PUNT: . 
\end{itemize}
}
\end{tcolorbox}

As for the \method method, we use the same prompt pattern as for encoder-only models:
\begin{tcolorbox}
    [colback=gray!20, colframe=gray!100, sharp corners, leftrule={3pt}, rightrule={0pt}, toprule={0pt}, bottomrule={0pt}, left={2pt}, right={2pt}, top={3pt}, bottom={3pt}]
{\small
\begin{itemize}
\item \textbf{Input text:} On the other hand, it looks pretty cool . The pos tag of On is: 
\item \textbf{Target text:} ADP
\end{itemize}
}
\end{tcolorbox}

\subsection{Dataset Statistics}
In \tabref{size} we show a basic statistic view of the PAN-X \citep{pan-etal-2017-cross} and UDPOS \citep{ud} datasets from the XTREME benchmark \citep{hu2020xtreme}. We use the original train-dev-test split from the datasets based one the XTREME benchmark. For training and validation, we use the English train and dev dataset, and for test we use the test sets of all languages. 

\begin{table}[ht]
  \centering
  \renewcommand\arraystretch{1.35}
  \small
  \begin{tabular}{lcccc}
    \toprule
\multirow{2}*{Task}   &  \multicolumn{3}{c}{Size} &  \multirow{2}*{\#Labels}   \\
\cmidrule(){2-4}
       & | Train |  & | Dev |  & | Test | &\\
\midrule
PAN-X & $20\ 000$ & $10\ 000$ & $10\ 000$ & 7\\
UDPOS & $21\ 253$  & $3\ 974$ & $3\ 973$ & 17\\

    \bottomrule
  \end{tabular}
\caption{Overview of the three datasets. Train and dev data size refers to the number of samples (sentences) for English. Test data size refers to the average number of samples for each target language.}
\tablabel{size}
\end{table}

\subsection{Tags and Meanings}
The tags of the two datasets and their detailed meanings are documented in \tabref{iob2tag} and \tabref{postag}.
\begin{table}[ht]
\renewcommand\arraystretch{1.15}
\setlength\tabcolsep{8pt}
\centering
\small
\begin{tabular}{lr}
\toprule
Tags   & Meaning                    \\
\midrule
B-LOC   & location (beginning)                  \\
B-ORG   & organization (beginning)                 \\
B-PER   & person (beginning)                     \\
I-LOC   & location (inside)                 \\
I-ORG & organization (inside)  \\
I-PER   & person (inside)                     \\
O   & other                 \\
\bottomrule
\end{tabular}
\caption{IOB2 tags}
\tablabel{iob2tag}
\end{table}

\begin{table}[ht]
\renewcommand\arraystretch{1.15}
\setlength\tabcolsep{8pt}
\centering
\small
\begin{tabular}{lr}
\toprule
Tags   & Meaning                    \\
\midrule
ADJ   & adjective                  \\
ADP   & adposition                 \\
ADV   & adverb                     \\
AUX   & auxiliary                  \\
CCONJ & coordinating conjunction   \\
DET   & determiner                 \\
INTJ  & interjection               \\
NOUN  & noun                       \\
NUM   & numeral                    \\
PART  & particle                   \\
PRON  & pronoun                    \\
PROPN & proper noun                \\
PUNCT & punctuation                \\
SCONJ & subordinating conjunction  \\
SYM   & symbol                     \\
VERB  & verb                       \\
X     & other                      \\
\bottomrule
\end{tabular}
\caption{Universal POS tags}
\tablabel{postag}
\end{table}

\subsection{Details of Experiments on MLLMs}
\seclabel{llm_details}
We apply our \method method to two MLLMs including:
\begin{itemize}
    \item BLOOMZ~\citep{muennighoff-etal-2023-crosslingual}, a multi-task fine-tuned version of the BLOOM~\citep{Scao2022BLOOMA1} model created by the BigScience community. We use \texttt{bloomz-7b1}, the version with 7.1 billion parameters, in our experiment.
    \item mT0~\citep{muennighoff-etal-2023-crosslingual}, a multi-task fine-tuned version of the mT5~\citep{xue-etal-2021-mt5} model. We use parameters \texttt{mt0-xxl}, the version with 13 billion, in our experiment.
\end{itemize}

In our exploratory study with MLLMs, we employ zero-shot English ICL without demonstrations~
\citep{DBLP:conf/iclr/ShiSF0SVCTRZ0W23}. The prompt template used for MLLMs is as follows:
\begin{tcolorbox}
    [colback=gray!20, colframe=gray!100, sharp corners, leftrule={3pt}, rightrule={0pt}, toprule={0pt}, bottomrule={0pt}, left={2pt}, right={2pt}, top={3pt}, bottom={3pt}]
{\small 
$T(X, x_i)=$\\
Named entity type: location organisation person place body name other \\
Sentence: $X$ \\
Named entity type of $x_i$ in the sentence is}
\end{tcolorbox}
The full experimental results of the two MLLMs on the PAN-X task are shown in \tabref{llm_results}.

\begin{table}[!htb]
\renewcommand\arraystretch{1.13}
\setlength\tabcolsep{10pt}
    \footnotesize
    \centering
\begin{tabular}{c | c c}
\toprule
\textbf{Lang.} & \textbf{bloomz-7b1} & \textbf{mt0-13b} \\
\midrule
\textbf{en} & 14.81 & 17.48 \\
\textbf{af} & 9.71 & 15.97 \\
\textbf{ar} & 20.63 & 19.37 \\
\textbf{az} & 10.00 & 17.25 \\
\textbf{bg} & 11.66 & 20.23 \\
\textbf{bn} & 23.27 & 26.68 \\
\textbf{de} & 10.96 & 15.32 \\
\textbf{el} & 8.70 & 14.11 \\
\textbf{es} & 17.20 & 20.70 \\
\textbf{et} & 12.16 & 18.12 \\
\textbf{eu} & 11.26 & 17.86 \\
\textbf{fa} & 19.51 & 20.08 \\
\textbf{fi} & 11.77 & 19.19 \\
\textbf{fr} & 20.55 & 20.11 \\
\textbf{gu} & 6.09 & 13.33 \\
\textbf{he} & 10.01 & 16.85 \\
\textbf{hi} & 17.98 & 23.10 \\
\textbf{hu} & 11.57 & 17.57 \\
\textbf{id} & 16.85 & 21.13 \\
\textbf{it} & 16.50 & 17.74 \\
\textbf{ja} & 7.58 & 4.79 \\
\textbf{jv} & 13.66 & 18.11 \\
\textbf{ka} & 8.11 & 18.23 \\
\textbf{kk} & 9.95 & 18.90 \\
\textbf{ko} & 11.32 & 19.10 \\
\textbf{lt} & 13.48 & 17.81 \\
\textbf{ml} & 13.37 & 22.53 \\
\textbf{mr} & 14.53 & 19.92 \\
\textbf{ms} & 22.08 & 19.10 \\
\textbf{my} & 3.37 & 18.59 \\
\textbf{nl} & 14.80 & 17.70 \\
\textbf{pa} & 12.74 & 17.14 \\
\textbf{pl} & 14.04 & 17.89 \\
\textbf{pt} & 19.35 & 20.46 \\
\textbf{qu} & 16.50 & 17.49 \\
\textbf{ro} & 21.19 & 20.18 \\
\textbf{ru} & 12.48 & 18.36 \\
\textbf{sw} & 17.02 & 23.72 \\
\textbf{ta} & 13.06 & 19.52 \\
\textbf{te} & 11.82 & 17.15 \\
\textbf{th} & 7.65 & 0.57 \\
\textbf{tl} & 25.10 & 22.20 \\
\textbf{tr} & 13.62 & 19.53 \\
\textbf{uk} & 11.74 & 19.69 \\
\textbf{ur} & 18.63 & 22.24 \\
\textbf{vi} & 16.86 & 17.05 \\
\textbf{yo} & 19.73 & 22.21 \\
\textbf{zh} & 6.86 & 5.25 \\
\midrule
\textbf{avg.} & \textbf{13.98} & \textbf{18.09} \\
\bottomrule
    \end{tabular}
    \caption{Full results of MLLMs on the PAN-X task.}
    \tablabel{llm_results}
\end{table}

\subsection{Detailed Results}
\seclabel{detailed_results}
We present the detailed results of the cross-lingual evaluation performance of Vanilla, Prompt Tuning, and \method in \tabref{panx_full} (PAN-X) and \tabref{udpos_full} (UDPOS).

\begin{table*}
  \centering
  \renewcommand\arraystretch{1.35}
  \setlength\tabcolsep{4pt}
  \footnotesize
  \begin{tabular}{lcccccccccccccc}
    \toprule
lang.       & en    & af    & ar    & az    & bg    & bn    & de    & el    & es    & et    & eu    & fa    & fi    \\
\midrule
B (Vanilla) & 83.83 & 78.07 & 44.09 & 67.58 & 78.31 & 70.09 & 79.10 & 71.85 & 73.95 & 77.96 & 65.44 & 42.43 & 78.74 \\
B (PT)      & 79.09 & 71.37 & 39.52 & 63.47 & 73.28 & 58.81 & 74.14 & 63.35 & 68.05 & 73.84 & 61.00 & 34.86 & 74.01 \\
B (\method)   & \textbf{92.80} & \textbf{90.87} & \textbf{62.62 }& \textbf{85.30} & \textbf{89.61} & \textbf{78.33} & \textbf{92.40} & \textbf{89.88} & \textbf{84.94} & \textbf{90.07} & \textbf{85.35} & \textbf{69.52} & \textbf{91.25} \\
\cmidrule(l){2-14}
X (Vanilla) & 81.31 & 75.03 & 47.26 & 61.37 & 77.02 & 68.97 & 74.07 & 74.93 & 70.51 & 70.73 & 58.07 & 48.73 & 75.44 \\
X (PT)      & 75.94 & 69.92 & 43.75 & 58.57 & 72.15 & 53.42 & 68.09 & 64.12 & 65.21 & 65.43 & 47.97 & 38.65 & 70.31 \\
X (\method)   & \textbf{92.21} & \textbf{90.02} & \textbf{67.84} & \textbf{84.02} & \textbf{88.20} & \textbf{72.06} & \textbf{91.22} & \textbf{91.22} &\textbf{ 83.63 }& \textbf{88.26} & \textbf{84.59} & \textbf{62.82} & \textbf{90.72} \\
\cmidrule(l){2-14}
T (Vanilla) & 77.14 & 76.94 & 49.99 & 62.00 & 72.98 & 60.32 & 76.19 & 76.88 & 67.81 & 74.25 & 67.12 & 40.46 & 75.93 \\
T (\method)   & \textbf{96.52} & \textbf{96.76} & \textbf{89.13} & \textbf{94.78} & \textbf{96.11} & \textbf{90.74} & \textbf{97.21} & \textbf{96.22} & \textbf{93.90} & \textbf{95.80} & \textbf{94.62} & \textbf{87.93} & \textbf{96.71} \\
\midrule
lang.       & fr    & gu    & he    & hi    & hu    & id    & it    & ja    & jv    & ka    & kk    & ko    & lt    \\
\midrule
B (Vanilla) & 80.40 & 53.89 & 55.80 & 68.17 & 76.16 & 61.21 & 81.10 & 28.25 & 61.58 & 67.94 & 47.21 & 61.60 & 74.41 \\
B (PT)      & 75.02 & 32.07 & 52.00 & 62.38 & 70.88 & 58.39 & 78.11 & 23.76 & 57.23 & 61.45 & 46.06 & 58.51 & 69.86 \\
B (\method)   & \textbf{87.15} & \textbf{87.22} & \textbf{83.27} & \textbf{80.88} & \textbf{90.91} & \textbf{77.99} & \textbf{91.24} & \textbf{69.29} & \textbf{80.28} & \textbf{87.25} & \textbf{80.95} & \textbf{83.94} & \textbf{87.99} \\
\cmidrule(l){2-14}
X (Vanilla) & 75.81 & 57.12 & 51.54 & 68.11 & 76.42 & 48.04 & 77.58 & 19.26 & 57.86 & 67.02 & 40.79 & 50.36 & 73.85 \\
X (PT)      & 69.14 & 47.54 & 43.64 & 60.58 & 70.17 & 45.33 & 71.55 & 16.98 & 41.49 & 57.22 & 40.66 & 44.73 & 67.08 \\
X (\method)   & \textbf{86.20} & \textbf{88.11} & \textbf{82.49} & \textbf{79.28} & \textbf{91.38} &\textbf{ 69.35} & \textbf{89.36} & \textbf{66.87} & \textbf{74.29} & \textbf{87.50} & \textbf{83.14} &\textbf{ 81.78 }& \textbf{88.09} \\
\cmidrule(l){2-14}
T (Vanilla) & 73.68 & 64.18 & 68.83 & 61.90 & 74.01 & 64.28 & 77.33 & 46.19 & 67.79 & 70.17 & 65.10 & 60.24 & 72.09 \\
T (\method)   & \textbf{94.55} & \textbf{96.17} & \textbf{92.93} & \textbf{92.69} & \textbf{96.98} & \textbf{91.22} & \textbf{96.35} & \textbf{89.71} & \textbf{90.59} & \textbf{96.02} & \textbf{93.73} &\textbf{ 93.40} & \textbf{95.57} \\
\midrule
lang.       & ml    & mr    & ms    & my    & nl    & pa    & pl    & pt    & qu    & ro    & ru    & sw    & ta    \\
\midrule
B (Vanilla) & 56.00 & 57.77 & 67.05 & 53.36 & 82.23 & 34.29 & 80.74 & 79.77 & 64.53 & 73.97 & 65.33 & 70.08 & 53.33 \\
B (PT)      & 50.35 & 51.17 & 63.17 & 43.18 & 77.78 & 31.36 & 77.38 & 74.00 & 46.06 & 59.57 & 58.14 & 60.57 & 49.08 \\
B (\method)   & \textbf{82.57} & \textbf{82.93} & \textbf{81.55} & \textbf{82.65} & \textbf{92.35} & \textbf{59.67} & \textbf{90.87} & \textbf{87.25} & \textbf{77.50} & \textbf{81.88} & \textbf{84.71} & \textbf{79.33} & \textbf{77.81} \\
\cmidrule(l){2-14}
X (Vanilla) & 59.85 & 60.74 & 66.13 & 53.41 & 79.67 & 50.31 & 77.64 & 76.83 & 60.49 & 70.45 & 62.54 & 69.51 & 54.62 \\
X (PT)      & 51.08 & 48.43 & 45.86 & 44.94 & 74.88 & 33.83 & 73.04 & 70.12 & 45.36 & 59.48 & 54.84 & 57.57 & 47.83 \\
X (\method)   & \textbf{85.55} & \textbf{81.75} &\textbf{ 74.39} & \textbf{85.10 }& \textbf{92.00} &\textbf{ 69.72 }& \textbf{90.66} & \textbf{85.99} & \textbf{77.57} & \textbf{83.60} & \textbf{80.65} & \textbf{77.32} & \textbf{81.30} \\
\cmidrule(l){2-14}
T (Vanilla) & 62.21 & 61.71 & 68.06 & 44.70 & 77.43 & 53.71 & 75.31 & 70.83 & 62.18 & 69.10 & 66.16 & 66.60 & 62.69 \\
T (\method)   & \textbf{94.77} & \textbf{93.42} & \textbf{85.70} & \textbf{93.66} & \textbf{96.93} & \textbf{86.18} & \textbf{96.34} & \textbf{94.81} & \textbf{87.35} & \textbf{94.16} &\textbf{ 94.07} & \textbf{91.90} & \textbf{92.52} \\
\midrule
lang.       & te    & th    & tl    & tr    & uk    & ur    & vi    & yo    & zh    & avg.  &       &       &       \\
\midrule
B (Vanilla) & 50.86 & 0.77  & 71.14 & 74.66 & 71.30 & 33.22 & 69.69 & 49.29 & 43.51 & 62.73 &       &       &       \\
B (PT)      & 47.77 & 0.54  & 71.54 & 67.16 & 65.20 & 26.49 & 67.17 & 37.71 & 40.73 & 56.76 &       &       &       \\
B (\method)   & \textbf{83.83} & \textbf{68.37} & \textbf{82.54} & \textbf{87.29} & \textbf{85.94} & \textbf{63.18} & \textbf{86.04} & \textbf{64.70} & \textbf{68.39} & \textbf{81.91} &       &       &       \\
\cmidrule(l){2-14}
X (Vanilla) & 48.20 & 3.09  & 69.84 & 75.58 & 73.43 & 59.48 & 67.92 & 50.25 & 25.28 & 61.30 &       &       &       \\
X (PT)      & 40.89 & 3.67  & 62.14 & 64.48 & 61.21 & 38.17 & 61.68 & 35.57 & 24.51 & 53.05 &       &       &       \\
X (\method)   & \textbf{84.73} & \textbf{19.56} & \textbf{78.35} & \textbf{89.35} & \textbf{85.74} & \textbf{61.11} & \textbf{82.18} & \textbf{66.38} & \textbf{66.09} & \textbf{80.03} &       &       &       \\
\cmidrule(l){2-14}
T (Vanilla) & 66.67 & 29.23 & 63.28 & 69.28 & 69.94 & 37.75 & 61.28 & 61.24 & 50.87 & 64.19 &       &       &       \\
T (\method)   & \textbf{94.82 }& \textbf{79.33} & \textbf{90.34} & \textbf{96.21} & \textbf{93.45} & \textbf{89.06} & \textbf{92.94} & \textbf{84.54} & \textbf{90.37 }& \textbf{92.82} &       &       &       \\
    \bottomrule
  \end{tabular}
  \caption{Detailed results of the cross-lingual evaluation on PAN-X.}
  \tablabel{panx_full}
\end{table*}

\begin{table*}

  \centering
  \renewcommand\arraystretch{1.35}
  \setlength\tabcolsep{4pt}
  \footnotesize
  \begin{tabular}{lcccccccccccccc}
    \toprule
lang.                      & en    & af    & ar    & bg    & de    & el    & es    & et    & eu    & fa    & fi    & fr    & he         \\
\midrule
B (Vanilla)                & 95.28 & 86.10 & 53.51 & 85.65 & 86.36 & 81.92 & \textbf{86.79} & 80.78 & 58.75 & 66.10 & 80.33 & \textbf{84.59} & 56.27      \\
B (PT)                     & 94.96 & 86.06 & 55.59 & 85.81 & 86.03 & 80.87 & 85.04 & 76.74 & 59.99 & 66.91 & 77.75 & 79.67 & 56.12      \\
B (\method) & \textbf{95.82} & \textbf{89.37} & \textbf{70.02} & \textbf{88.45} & \textbf{89.46} & \textbf{85.72} & 85.93 & \textbf{84.64} & \textbf{68.86} & \textbf{68.33} & \textbf{82.96} & 84.43 & \textbf{80.68}      \\
\cmidrule(l){2-14}
X (Vanilla)                & 95.64 & 87.88 & 65.41 & 88.48 & 88.03 & 86.63 & \textbf{88.31} & 85.96 & 70.07 & 69.22 & 85.32 & \textbf{86.57} & 66.38      \\
X (PT)                     & 95.18 & 87.92 & 65.69 & 88.35 & 87.76 & \textbf{86.78} & 87.98 & 84.96 & 66.71 & 68.57 & 84.60 & 86.21 & 66.12      \\
X (\method) & \textbf{96.05} & \textbf{89.88} & \textbf{70.06} & \textbf{89.04} & \textbf{89.61} & 86.14 & 87.08 & \textbf{86.90} & \textbf{71.95} & \textbf{70.04} & \textbf{85.80} & 81.21 & \textbf{80.50}      \\
\cmidrule(l){2-14}

T (Vanilla)                & 89.67 & 85.02 & 63.56 & 78.38 & 79.86 & 75.44 & 83.99 & 78.35 & 68.49 & 66.47 & 77.50 & 82.10 & 64.19      \\
T (\method) & \textbf{97.57} & \textbf{92.18} & \textbf{78.79} & \textbf{92.72} & \textbf{92.35} & \textbf{88.50} & \textbf{89.72} &\textbf{ 89.93} & \textbf{82.63} & \textbf{81.59} & \textbf{89.18} & \textbf{90.51} & \textbf{87.87}      \\
\midrule
lang.                           & hi    & hu    & id    & it    & ja    & kk    & ko    & lt    & mr    & nl    & pl    & pt    & ro         \\
\midrule
B (Vanilla)                & 63.54 & 78.92 & 71.67 & \textbf{88.46} & 47.05 & 70.53 & 50.82 & 79.79 & 70.35 & 89.00 & 81.77 &\textbf{ 86.44} & 78.00      \\
B (PT)                     & 64.54 & 78.40 & 71.47 & 86.78 & 46.77 & 70.19 & 51.63 & 76.93 & 67.24 & 88.49 & 81.15 & 86.02 & 77.14      \\
B (\method) & \textbf{73.16} & \textbf{79.48} &\textbf{ 76.30} & 86.45 & \textbf{52.10} & \textbf{74.98} & \textbf{64.68} & \textbf{83.54} & \textbf{75.75} & \textbf{89.50} &\textbf{ 84.97 }& 85.36 & \textbf{81.15}      \\
\cmidrule(l){2-14}

X (Vanilla)                & 69.35 & \textbf{82.97} & 72.83 & 87.79 & 25.62 & 76.14 & 52.75 & 84.67 & 82.61 & 89.26 & 83.91 & \textbf{87.16} & 84.23      \\
X (PT)                     & 69.19 & 82.72 & 72.50 & \textbf{88.88} & 22.17 & 74.93 & 53.29 & 83.11 & 81.22 & 88.95 & 84.24 & 87.11 & 83.80      \\
X (\method) & \textbf{72.98} & 80.90 & \textbf{76.93} & 86.53 & \textbf{54.78} & \textbf{76.61} & \textbf{64.14} & \textbf{87.16} & \textbf{80.09} & \textbf{89.54} & \textbf{85.74} & 86.37 & \textbf{85.69}      \\
\cmidrule(l){2-14}

T (Vanilla)                & 69.21 & 76.85 & 72.11 & 82.71 & 50.81 & 71.57 & 51.22 & 76.92 & 72.58 & 83.85 & 77.39 & 82.78 & 74.51      \\
T (\method) &\textbf{ 87.89} & \textbf{90.75} & \textbf{85.92} & \textbf{90.99} & \textbf{78.12} & \textbf{87.18} & \textbf{76.79} & \textbf{89.73} & \textbf{89.76} & \textbf{93.01} & \textbf{91.16} & \textbf{90.74} & \textbf{88.58}      \\
\midrule
lang.                           & ru    & ta    & te    & th    & tl    & tr    & uk    & ur    & vi    & wo    & yo    & zh    & avg. \\
\midrule
B (Vanilla)                & 85.82 & 58.78 & \textbf{76.63} & 41.08 & \textbf{82.30} & 69.46 & 81.04 & 55.04 & 55.98 & 30.93 & 59.56 & 62.62 & 70.89      \\
B (PT)                     & 86.58 & 59.33 & 74.25 & 37.00 & 77.59 & 66.17 & 81.32 & 56.01 & 54.69 & 29.19 & 57.50 & 63.88 & 69.91      \\
B (\method) & \textbf{90.02} & \textbf{72.21} & 75.70 &\textbf{ 56.92} & 81.71 & \textbf{71.29} & \textbf{86.95} & \textbf{67.08} &\textbf{ 58.77 }& \textbf{33.38} & \textbf{65.29} & \textbf{72.17} & \textbf{76.16}      \\
\cmidrule(l){2-14}

X (Vanilla)                & 89.16 & 61.94 & \textbf{84.38} & 44.73 & 86.80 & \textbf{74.22} & 85.22 & 58.88 & 58.48 & \textbf{30.07} &\textbf{ 26.12} & 32.08 & 72.42      \\
X (PT)                     & 88.50 & 61.91 & 82.11 & 40.80 & \textbf{88.64} & 72.74 & 84.85 & 60.68 & 57.34 & 28.79 & 25.07 & 33.81 & 71.86      \\
X (\method) & \textbf{90.70} & \textbf{72.78} & 83.79 &\textbf{ 70.01} & 82.11 & 73.85 & \textbf{87.17} & \textbf{66.82} & \textbf{59.79} & 19.38 & 19.53 & \textbf{76.38} & \textbf{76.16}      \\
\cmidrule(l){2-14}

T (Vanilla)                & 82.12 & 62.85 & 78.65 & 64.06 & 73.73 & 68.58 & 77.17 & 64.63 & 58.43 &\textbf{ 54.89} & 66.74 & 43.90 & 71.39      \\
T (\method) & \textbf{93.48} &\textbf{ 83.23} & \textbf{90.65} & \textbf{79.67 }&\textbf{ 93.11} & \textbf{85.46} & \textbf{90.67} & \textbf{85.10} & \textbf{79.03} & 54.01 &\textbf{ 72.71} & \textbf{82.43} & \textbf{86.11 }      \\
    \bottomrule
  \end{tabular}
  \caption{Detailed results of the cross-lingual evaluation on UDPOS.}
  \tablabel{udpos_full}
\end{table*}

\end{CJK}
\end{document}